\documentclass[sigconf, nonacm]{acmart}
\usepackage{fancyhdr}
\renewcommand\footnotetextcopyrightpermission[1]{} % removes footnote with conference information in first column
\usepackage{amsmath}

\usepackage{multirow}
\usepackage{array}
\usepackage{epsfig}
\usepackage{subfig}
\usepackage{graphicx}  %插入图片的宏包
\usepackage{float}  %设置图片浮动位置的宏包
\usepackage{multirow}
\AtBeginDocument{%
  \providecommand\BibTeX{{%
    \normalfont B\kern-0.5em{\scshape i\kern-0.25em b}\kern-0.8em\TeX}}}

\setcopyright{acmcopyright}
\copyrightyear{2020}
\acmYear{2020}
\acmDOI{10.xxxx/xxxxxxx.xxxxxxx}

\begin{document}

\title{Impact of Medical Data Imprecision on
  Learning Results
  % How the error margin effect the disease progress prediction model
}

\author{Mei Wang}
\authornote{A part of work is done while visiting UCSB.}
\author{Haiqin Lu}
%\email{}
\affiliation{%
  \institution{Donghua University, China}
 % \streetaddress{P.O. Box 1212}
}
\email{wangmei@dhu.edu.cn}

\author{Jianwen Su}
\affiliation{
  \institution{Univ of California, Santa Barbara, USA}
%  \streetaddress{1 Th{\o}rv{\"a}ld Circle}
}
\email{su@cs.ucsb.edu}

\renewcommand{\shortauthors}{}

\begin{abstract}
  Test data measured by medical instruments often carry
  imprecise ranges
  that include the {\em true} values.
  The latter are not obtainable in virtually all cases.
  Most learning algorithms,
  however,
  carry out arithmetical calculations that are
  subject to uncertain influence in both the learning process
  to obtain models and applications of the learned models
  in, e.g. prediction. % and classifications.
  In this paper,
  we initiate a study on the impact of imprecision on
  prediction results in a healthcare application
  where a pre-trained model is used
  to predict future state of hyperthyroidism for patients.
  We formulate a model for data imprecisions.
  Using parameters to control the degree of imprecision,
  imprecise samples for comparison experiments
  can be generated using this model.
  Further,
  a group of measures are defined to
  evaluate the different impacts quantitatively.
  More specifically,
  the statistics to measure the inconsistent prediction
  for individual patients are defined.
  We perform experimental evaluations to 
  compare prediction results based on the data
  from the original dataset and
  the corresponding ones generated from
  the proposed precision model using
  the long-short-term memories (LSTM) network.
  The results against a real world hyperthyroidism dataset
  provide insights into how small imprecisions
  can cause large ranges of predicted results,
  which could cause mis-labeling and
  inappropriate actions (treatments or no treatments)
  for individual patients.
\end{abstract}


%\ccsdesc[300]{Machine learning}
%\ccsdesc{Machine learning approaches}


\maketitle

\section{Introduction}
Clinical lab tests play an increasingly important role
in today's healthcare.
From early detection of diseases
to diagnosis to personalized treatment programs,
lab tests guide more than 70\% of medical decisions
and personalized medication \cite{labtesting14}.
The availability of medical and healthcare datasets
makes healthcare one of the focused areas
of applying machine learning techniques
in order to improve healthcare \cite{GRAM2017,KAME2018}.
However,
due to limitations of equipment, instruments, materials,
test methods, etc.,
data inaccuracy always occurs,
leading to uncertainty in measured results.
It becomes an interesting problem to understand the impact of
accuracy of clinical laboratory test results
on effectiveness of machine learning.
This paper initiates a study to quantify
the impact of medical data imprecision
on LSTM \cite{LSTM02} prediction results.

Imprecision of clinical lab test results is often due to
systematic and random factors.
Systematic factors are mostly reproducible and
tend to skew results consistently in the same direction.
For example,
the attenuation of the light source of the instrument
will cause
test results to shift to one side.
Random factors are unpredictable during operations,
examples include
expired reagents, 
expired controls or calibrators,
or failure in sampling system. 
Since 
random factors cannot be easily attributed 
to certain reasons, 
it is difficult to eliminate them.
Usually large imprecision margins 
are unallowable for lab tests.
Therefore, 
each lab follows some specified quality control 
and process control protocols 
to ensure 
that the test results 
are within respective tolerable ranges 
(or imprecision ranges, 
values in this range are acceptable, 
though imprecise). 
Such ranges 
for common biochemical tests 
are provided in \cite{CV14}. 

Earlier studies focused on
how the noise 
(erroneous values, 
missing or incomplete values)
in datasets
would impact the model learning and the way
to deal with them \cite{noisy2018, Noiselabel, Yeo2019}. 
However, 
it is not clear 
what effect
imprecision ranges have
in prediction results 
when a pre-trained model 
is used to 
make prediction for new sample.
Based on a dataset 
in the study of hyperthyroidism progress prediction, 
we study in this paper
the impact of imprecision 
on prediction results.
%when the prediction model is in use.  
We first formulate a model 
to represent data imprecision 
with a parameter to control the degree of imprecision. 
Imprecise samples for comparison experiments 
can be generated using this model. 
A group of measures are then defined to 
evaluate the different impact 
in a quantitative way. 
More specifically, 
the statistics to measure the inconsistent prediction 
for individual patients is defined. 
We carried out 
a set of experiments 
to compare the prediction results
based on the data 
from the medical database 
and its corresponding ones 
generated from the proposed imprecision model. 
The experiments are conducted 
against a real world dataset. 
The experimental results
provide fresh insights into
the reasons
that the small imprecise 
could cause the inconsistent prediction results 
for individual patients, and
lead to unreliable results. 

% The main contributions of this paper are as follows:
% \begin{itemize}
% \item
% \item
% \item
% \end{itemize}

The %remainder of the
paper is organized as follows.
Section 2
motivates the study  with a real application example.
Section 3
introduces an imprecision model and measurements.
Section 4
presents experimental results.
Section 5 concludes the paper.
% and discusses future work.

\vspace{-1.5mm}

\section{Motivations}

Test data measured by medical instruments often carry imprecise ranges
that include the {\em true} values
that are not obtainable in almost all cases.
Most learning algorithms,
however,
carry out arithmetical calculations in both the learning process
to obtain models and applications of the learned models
in, e.g. prediction.
In this section,
we present an application example and
illustrate how the imprecision could cause
prediction results to be wrong.

The application occurs
%carried out 
in a collaboration 
between researchers in Donghua University (Shanghai, China) and
physicians in Ruijing Hospital (Shanghai, China).
The aim is to develop a personalized prediction service
to identify hyperthyroidism progress for patients.

Patients with hyperthyroidism (a chronic disease)
always experience stages of 
occurrence, remission, cure or recurrence. 
Hyperthyroidism occurs 
when the thyroid gland produces too much thyroxine (hormone).
Lab tests are an integral part of
the diagnosis process
to assess the progressing stage of hyperthyroidism.
Typical thyroid tests 
include blood tests for thyroid-stimulating hormone (TSH),
free thyroxine and TSH receptor antibodies (TRAb). 
Low TSH levels may be a sign of
an overactive thyroid (hyperthyroidism).

For a patient with 
a period of treatment, 
TSH may become normal.
After stopping treatment for a while,
if the TSH level becomes low again,
it could mean a recurrence of hyperthyroidism.
If such recurrence is predicted in advance
for a patient with a high probability,
additional treatments
can be carried out early,
such as the special medication strategy, 
unconventional inspection, 
or ablative therapies 
including radioactive iodine treatment (RAI) 
or the surgical removal of the thyroid tissue.
In the collaboration, 
a prediction model was developed \cite{Donghuawork}
%in the application 
to predict the progress of hyperthyroidism 
two years in advance
based on the feature data 
in the first six months. 
The feature data includes the patient's basic information
such as gender, age
as well as the value of the key tests FT3, FT4, TSH, and TRAb. 

Accuracy of clinical lab test results is fundamental.
Under ideal circumstances,
none of systematic and random factors occur,
but this is not achievable 
in daily practice.
Thus the test results are
given allowable imprecision ranges.
For example,
the desirable and optimum levels of allowable
imprecision specifications for TSH are
9.65\% and 4.825\% resp. \cite{allowableimp}
On the other hand,
even test results fall in small ranges,
the imprecise values may cause large range of predicted results,
which may further cause mis-labeling and wrong actions
(treatments or no treatments).
This is best explained with a representative clinical case
presented below.
% (illustrated in Fig.\,\ref{fig:example}).

\begin{figure}\centering
\includegraphics[width=8.5cm]{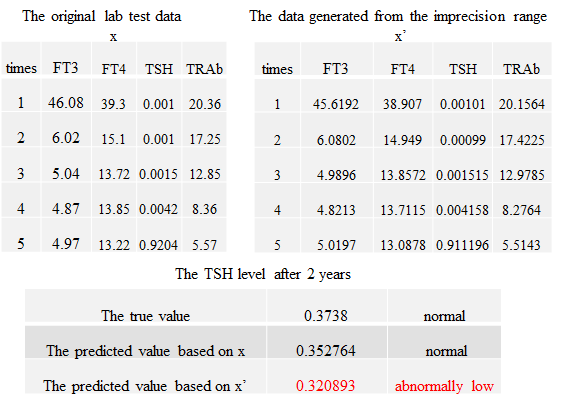}
\vspace*{-6mm}  
\caption{Test data, shifted data, and prediction results}
%Prediction results of the given patient case.
\label{fig:example}
\vspace*{-4mm}
\end{figure}

Consider a patient from the dataset studied in the collaboration
mentioned in the above.
Patient X (female, 34 years old)
first visited the hospital with a low TSH level.
The trained prediction model is used
to learn whether she has a high risk of recurrence.
The top left table in Fig.\,\ref{fig:example}
shows 
the 5 lab test results in the first 6 months.
Based on these original lab test data
the prediction result is shown
in the second line of the bottom table in
Fig.\,\ref{fig:example}:
the predicted 
TSH level was at $0.352764$ and within the normal range.

To find out how imprecision would impact
prediction results
we introduce a small margin of $\pm1\%$
to the original lab test data.
The top right table of Fig.\,\ref{fig:example}
lists the slightly shifted data with the imprecision interval.
% However, 
% since there is the imprecision problem, 
% we obtain another set of data 
% (illustrated in the top right of Figure \ref{fig:example}). 
Using this set of shifted data,
the prediction model would forecast
patient X's TSH level at $0.320893$,
which falls into the ``abnormally low'' rage.
This abnormally low TSH level
indicates a high probability of recurrence for the patient.
Had this prediction result is used,
the suggested diagnosis 
and treatment would be different.
The bottom table in Fig.\,\ref{fig:example}
illustrates the actual TSH level of $0.3738$ after 2 years (Line 1),
the predicted value based on the original test data (Line 2),
and predicted value based on shifted data.

In practice,
physicians are usually aware of
measuring errors in test data
in the process of diagnosis and treatment.
They typically
do not make their judgement solely relying on laboratory tests.
In addition to laboratory examinations,
there are also imaging, pathology, device examinations, and
medical history information.
In addition,
physicians have prior background knowledge of
clinical lab tests' imprecise results.
A variety of information sources and background
can be used to reduce the influence of lab test result imprecisions
on diagnostic results.
Overall lab test result imprecisions usually have less
clinical impact in the current practice.
For the prediction models obtained from machine learning,
however,
it would be almost impossible 
to obtain most of the above data in the learning/prediction process.
% the model can be robust to errors, but
It is critical to avoid
prediction mistakes due to imprecision as much as possible.
% very difficult to obtain the above complete information.
% At this time, even for the well trained model, a small errors in
% laboratory test results could influence the outcome of the prediction
% model.
% So it is important to understand the impact of those error margin on
% behavior of the model.

\section{\hspace*{-5pt}Imprecision Model and Measurement}

To quantitatively measure the impact of data imprecision
on prediction results,
in this section we formulate a model to represent data imprecision.
The model is agnostic with respect to
the actual values and thus allows to focus
on prediction accuracy.

In real medical applications (databases),
for a given test measurement,
we can only get one certain test result value.
Our goal is to understand the impact of imprecision margin on behavior of
the prediction model.
For the obtained value $x$,
we define $x'$ that has a small difference with $x$ as follows:
\begin{equation}
  x' = x \pm \Delta_x * x
\label{eqn:x'}
\end{equation}
where $\Delta_x = \frac{x'-x}{x}$ is used to control 
the variation of $x$. 

Let $f$ be a prediction function.
Then the predicted value $y' = f(x')$
is calculated from $x'$.
% through functional relationship $f$,
Typically,
imprecision in the input variables will propagate through
the calculation to an imprecision in the output $y'$.
To numerically measure the degree of $\Delta_x$ affecting $y$, we define:
\begin{equation}
f(x') = f(x) \pm \Delta_y * f(x)
\end{equation}
where $\Delta_y = \frac{f(x') - f(x)}{f(x)}$ measures the change of
$y$ due to $\Delta_x$.

In general measuring changes of $\Delta_y$ as $\Delta_x$ varies
provides a fairly informative indication of imprecision propagation.
For the dataset and prediction in our study,
we would also like to measure the mis-labelings due to the imprecise
prediction results.
% Next,
% we define the measures to evaluate how $\Delta_x$ impact the label of
% predicted value.
After predicting $f(x)$,
the label $l(f(x))$ can be obtained by
comparing the value of $f(x)$ with the reference range.
For example,
we know the normal range of the blood test measurement TSH is
$[0.34, 5.6]$ (mIU/L).
A predicted value of 5.8458 indicates abnormal high label.
Assume a group of $n$ tested patients forms the test dataset
$D_{\textrm{Test}} = (x_1, x_2, \ldots, x_n)$.
% where $n$ denotes the number of the test samples.
Since $l(f(x))$ is a classification label,
accuracy is used to measure the prediction performance.
Given one label $l$, let $\textit{TP}$
model the positive samples in the test set and
the predictions are correct.
$\textit{TN}$ represents the negative samples and the predictions
are correct.
Accuracy is then defined as follows:
$$
\textit{Accuracy}  =  \frac{|\textit{TP}\,| + |\textit{TN}\,|}{|D_{\textrm{Test}}|}
= \frac{|\textit{TP}\,| + |\textit{TN}\,|}{n}
$$

Similarly,
for each $x_i$,
we generate the corresponding $x_i'$ according to
Eq.\,(\ref{eqn:x'}).
The $x_i'$'s samples form $D_{\textrm{Test}}'$.
For prediction model $f$ and
test dataset $D_{\textrm{Test}}'$,
we can calculate $\textit{TP}'$ and $\textit{TN}'$,
then we have:
$$
\textit{Accuracy}' =
\frac{|\textit{TP}'| + |\textit{TN}'|}{|D_{\textrm{Test}}'|}
= \frac{|\textit{TP}'| + |\textit{TN}'|}{n}
$$
We further define:
$$
\begin{array}{rcl}
  \Delta_{\textit{Accuracy}}  & = & \textit{Accuracy}' - \textit{Accuracy} \\
  & = &\displaystyle
  \frac{|\textit{TP}'| + |\textit{TN}'| - (|\textit{TP}\,|+|\textit{TN}\,|)}{n}
\end{array}
$$

We also define the following two patient sets:
$$
\begin{array}{l}
N-P =  (\textit{TP}' \cup \textit{TN}') - (\textit{TP} \cup \textit{TN}) \\ 
P-N =  (\textit{TP} \cup \textit{TN}) - (\textit{TP}' \cup \textit{TN}')
\end{array}
$$
Finally, we obtain:
$$
\begin{array}{r@{~}c@{~}l}
  \textit{Count\_gain} & = & |(N-P)| - |(P-N)|\\
  \textit{Count\_inconsistent}  & = & |(N-P)| + |(P-N)|
\end{array}
$$
Here $N-P$ denotes the patients
whose predicted label on $x$ are wrong,
while the predicted label on $x'$ are correct.
$P-N$ denotes the patients whose predicted label on $x$ are correct,
while the predicted labels on $x'$ are wrong.
$\textit{Count\_gain}$ denotes the positive gain.
Since
$\textit{TP} \cap \textit{TN} = \emptyset$ and
$\textit{TP}' \cap \textit{TN}' = \emptyset$,
we have $\Delta_{\textit{Accuracy}} = count\_gain / n$.
We can see $\Delta_{\textit{Accuracy}}$
only counts the difference between $N-P$ and $P-N$.
While in real application,
we need to pay attention to every patient whose predicted result
changes.
Thus we define $\textit{Count\_inconsistent}$ which denotes number of
the inconsistent label prediction.
This provides new insight into a disagreement between predictions for individual patients.

\begin{figure*}[htbp]
\centering
\begin{minipage}[t]{0.48\textwidth}\centering
\includegraphics[width=7.8cm]{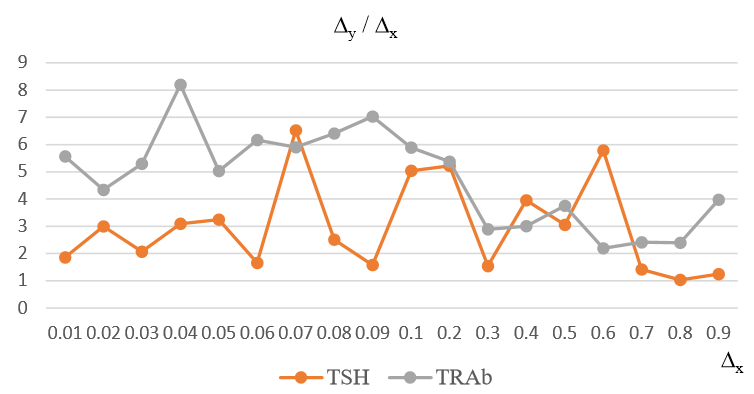}
\vspace*{-2.5mm}  
\caption{The change trend of $\Delta_y$ with $\Delta_x$}
\label{fig:deltay}
\vspace*{-2mm}
\end{minipage}
\begin{minipage}[t]{0.48\textwidth}\centering
\includegraphics[width=7.8cm]{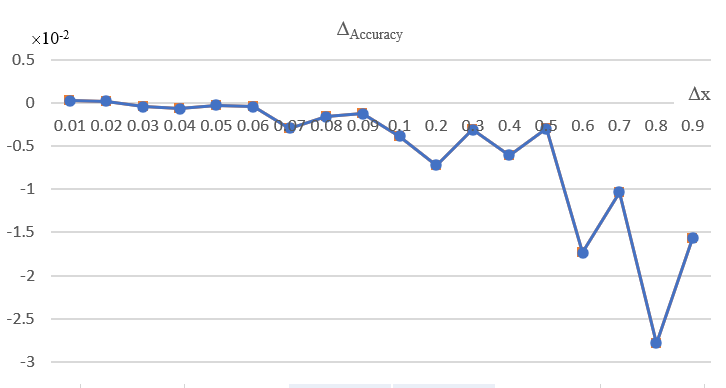}
\vspace*{-2mm}  
\caption{The change trend of $\Delta_{Accuracy}$}
\label{fig:accuracy}
\end{minipage}
\quad
\begin{minipage}[t]{0.48\textwidth}
\centering
  \includegraphics[width=7.8cm]{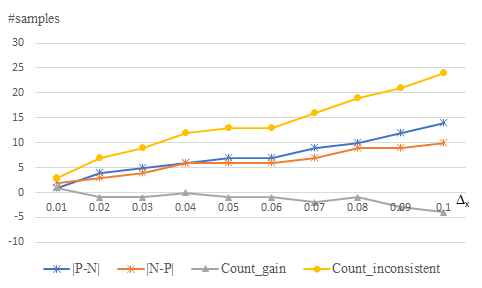}
\vspace*{-4mm}  
  \caption{The change trend of $Count\_inconsistent$}
  \label{fig:count}
\end{minipage}
\begin{minipage}[t]{0.48\textwidth}
\centering
  \includegraphics[width=7.8cm]{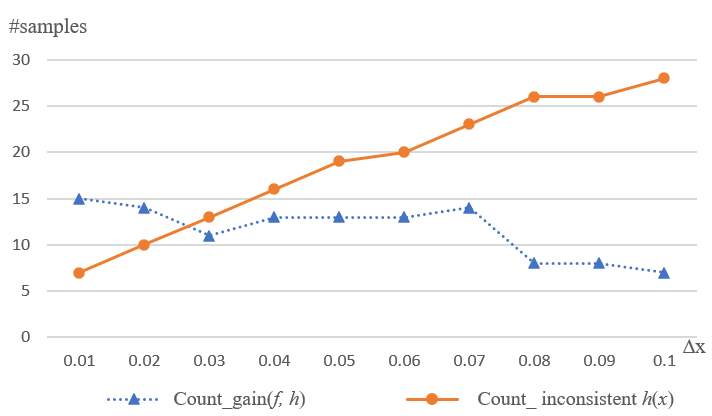}
\vspace*{-4mm}  
  \caption{The performance of the model with supplemental training set}
  \label{fig:compare}
\end{minipage}
\vspace*{-3mm}
\end{figure*}

\section{Experimental Evaluations}

To measure how imprecision margins effect prediction results,
we run experiments against
a large real-world clinic hyperthyroidism dataset.
We exploit LSTM \cite{LSTM02}
to predict two measurements closely related with
hyperthyroidism in future two years
based on the test data in the first six months.
TSH and TRAb are used as predicted targets.

\subsection{Dataset}

Our hyperthyroidism dataset is generated from Ruijin hospital, 
a reputable hospital in Shanghai, China,
including nearly 10 years of patient records.
The patients who satisfy the following two conditions
are selected to the dataset.
The first condition is that the first three diagnoses are
about the hyperthyroidism
and at least one of the of TRAb levels
in the first three test results is abnormally high.
This condition is used to filter out the
pseudo-hyperthyroidism.
The second condition is that the time periods of records
are more than 2 years because our method
is designed for a future prediction.
There are 2,460 patients in the final dataset
used in our experiments.

% It should be noted that
In the dataset patients
have different numbers of measurement records.
According to the statistics of the records,
we found that almost half of the patients
have about 7 records in the first six months.
So we fix the number of the examination results
at the first 6 month to be 7.
For the patients whose records were less than 7 times, we
filled missing ones with the test results closest in time.

After preprocessing the dataset,
we divided the whole dataset into a training set and a test set.
The training set consists of 1960 patients and
the remaining 500 patients are in the test set.

In the LSTM network training,
we use Truncated Normal Distribution to initialize
the weights of the input and output layers.
The hidden layer is expanded to 2 layers,
each layer containing 128 LSTM cell units.
We employ dropout method to reduce over-fitting and
apply the Adam-Optimizer in training.
Each experiment runs for 10 times and each data given
in the experimental results is the average of the 10 runs.
\vspace*{-4mm}

\subsection{Results}

In experiments,
we study the change trend of $\Delta_y$,
$\Delta_{\textit{Accuracy}}$, and
$\textit{Count\_inconsistent}$ with $\Delta_x$.
The prediction model is trained based on the original training data.

Fig.\,\ref{fig:deltay} illustrates
the ratio $\Delta_y / \Delta_x$ when $\Delta_x$ increases.
From the figure,
we can see that generally the ratio is larger than 1,
and that $\Delta_y$ grows faster than $\Delta_x$.
When $\Delta_x = 0.04$, the $\Delta_y$ for TRAb expands more than 8 times.
It clearly show that a slight change in $x$
may cause a large change in $y$.
When $\Delta_x$ is large,
it seems that 
the ratio shows a slight downward trend.
However,
when $ \Delta_x$ is large, $\Delta_y$ will be even larger.

Fig.\,\ref{fig:accuracy} illustrates the change trend of accuracy
when $\Delta_x$ increases.
It can be observed that
when $\Delta_x$ is small,
there is little change in accuracy.
Fig.\,\ref{fig:accuracy} indicates that
when $\Delta_x$ is greater than 0.1,
the accuracy begins to decline obviously.
Also we could see
that the decline is not stable.
We observed the results 
and found 
that 
since the normal ranges of TSH 
and the abnormal lower range of TRAb 
is much smaller
compared with their other ranges. 
The prediction label 
could easily
change from correct to wrong
or from wrong to correct 
for these ranges
by introducing the imprecision to the data, 
leading to the unstable decline. 

On the other hand, 
although accuracy did not decrease significantly 
when $\Delta_x$ is small,
we could not ignore the impact of the small $\Delta_x$
on the prediction results. 
The prediction results 
might be unreliable 
for individual patient. 
It is demonstrated 
in Fig.\,\ref{fig:count}.
We provide the value of $|N-P|, |P-N|, \textit{Count\_gain}$,
and $\textit{Count\_inconsistent}$
when $\Delta_x$ takes different values in Fig.\,\ref{fig:count}.
From the figure,
we can see that when $\Delta_x = 0.05$,
$\textit{Count\_inconsistent} = 13$, which means that
13 patients' prediction results have changed.
$\textit{Count\_inconsistent}$
provides a new way to evaluate the performance of the prediction model.

Finally,
we also tried out an idea of using a training set with imprecision,
hoping to learn a model that is less sensitive to imprecision.
Specifically,
for each $x$ we generate sample
$x_1' = x + \Delta_x^T * x$; $x_2' = x - \Delta_x^T * x$.
Then we randomly choose $x_1'$ or $x_2'$ as $x'$.
Gather all of the $x'$s to form the supplementary data set.
We combine the original data set with the supplemental data set to
form the new dataset,
then retrain the model based on the new dataset,
and finally obtain the new prediction model $h(x)$. 
At first, 
We count the number 
of two patient sets $N-P$ and $P-N$ 
according to 
the prediction results of $h(x)$ and $f(x)$, 
and then 
obtain $\textit{Count\_gain}(f,h)$, 
which is illustrated in Fig.\,\ref{fig:compare}.
From the figure, we can infer that by comparing the accuracy,
$h(x)$ definitely outperforms baseline prediction model,
since $\textit{Count\_gain}(f,h)$ 
is significantly greater than 1 
especially when $\delta x$ is small.

Also we use $h(x)$ to predict the samples in imprecision ranges 
and then calculate $\textit{Count\_inconsistent}$ 
based on the original test set 
and the imprecise one.
Fig.\,\ref{fig:compare} shows
that the $\textit{Count\_inconsistent}$ of $h(x)$ 
is also very large.
When $\Delta_x = 0.05$,
$\textit{Count\_inconsistent} = 19$, which is even larger than that of $f(x)$. 
In $h(x)$ learning process, 
more data are involved 
in training process, 
it can be inferred 
that more delicate curve 
that fits the data were learned. 
Correspondingly, 
the global accuracy is improved. 
However, the model structure of 
$h(x)$ is same as $f(x)$, which did not 
deal with the imprecise problem, so the  $\textit{Count\_inconsistent}$ of $h(x)$ has not decreased.

\section{Conclusion}

Medical data imprecision is
a common issue that could be easily ignored.
This paper carried out a set of experiments
to understand the impact of imprecision
on prediction results in the applications
of hyperthyroidism progress prediction.
The study has direct guidance on practical healthcare applications.
In addition, it motivates to build robust models
that can take imprecisions into account 
with better generalization.
Much work remains, including to understand and quantify
the impact of imprecision in more general settings
on learning models including specifically
medical datasets.
The general principles that
guide the impact are also interesting to explore and develop;
Based on such principles,
more effective methods to deal with
data imprecision are to be investigated.

\begin{acks}
We are very grateful to
Qiuya Lu 
% (Department of Clinical Laboratory Medicine, 
(Ruijin Hospital, Shanghai)
% Shanghai Jiao Tong University 
% School of Medicine) 
for the discussions on imprecision of data in test results.
A part of work by Wang was supported
by 
the National Key R\&D Program 
of
China under Grant 2019YFE0190500.
\end{acks}
%\begin{acks}
%To Robert, for the bagels and explaining CMYK and color spaces.
%\end{acks}

\bibliographystyle{ACM-Reference-Format}
\bibliography{MaggieRef}

\end{document}